%% file: main.tex
\documentclass[letterpaper]{article}

\usepackage{natbib,alifeconf}  
\usepackage[utf8]{inputenc} 
\usepackage[T1]{fontenc}    
\usepackage{url,hyperref}
\usepackage{booktabs}       
\usepackage{amsfonts}       
\usepackage{nicefrac}       
\usepackage{amsmath}
\usepackage{cleveref}
\crefname{figure}{Figure}{Figures}
\Crefname{figure}{Figure}{Figures}
\usepackage{amssymb}
\usepackage{tikz}
\usepackage{subcaption}
\usetikzlibrary{positioning, arrows.meta, shapes.geometric, calc, fit, backgrounds, shapes, decorations.pathmorphing, decorations.pathreplacing, backgrounds, shadows, patterns}

\newcommand\blfootnote[1]{%
  \begingroup
  \renewcommand\thefootnote{}\footnote{#1}%
  \addtocounter{footnote}{-1}%
  \endgroup
}

\title{Self-Organising Digital Circuits}

\author{
    Marcello Barylli$^{1}$\thanks{\hspace{0.5em}Denotes equal contribution.},
    Gabriel B\'{e}na$^{2}$\footnotemark[1],
    Alexander Mordvintsev$^{3}$,
    Eleni Nisioti$^{1}$, \and
    Sebastian Risi$^{1,4}$\\
    \mbox{}\\
    $^1$IT University of Copenhagen, Denmark\\
    $^2$Imperial College London, United Kingdom\\
    $^3$Google, Paradigms of Intelligence Team, Zurich, Switzerland\\
    $^4$Sakana AI, Tokyo, Japan\\
    \small{\texttt{bary@itu.dk, g.bena21@imperial.ac.uk}}
}

\begin{document}

\maketitle

\begin{abstract}
Fault tolerance in classical computing has traditionally relied on static strategies like hardware redundancy and error-correcting codes. Biological systems, in contrast, exhibit adaptive plasticity, maintaining function through dynamic re-organisation around damage. Inspired by this principle, we introduce Self-Organising Digital Circuits, framing functional logic generation and maintenance as a meta-learning problem on graphs. Our architecture employs a topology-masked Transformer that configures the Lookup Tables (LUT) of a circuit's Boolean gates. Extending the pattern-generation paradigm of Neural Cellular Automata, it navigates the degenerate Boolean search space to satisfy a computational task, rather than regenerating a fixed target state. We demonstrate that it can self-assemble functional circuits from scratch and rapidly re-route logic around permanent, unseen hardware faults. For soft errors, the policy achieves near-perfect recovery (>99.99\% accuracy) from damage sizes far exceeding training conditions. We further observe generalisation across circuit scales: accuracy improves on graphs substantially wider than those seen during training. This work bridges the principles of biological self-organisation with the practical domain of digital hardware.

\end{abstract}

Data/Code + \textbf{Circuit Visualization} available at: \url{https://github.com/GabrielBena/boolean_nca_cc/tree/gabi}
\blfootnote{\textcopyright\ 2026 Marcello Barylli, Gabriel B\'{e}na, Alexander Mordvintsev, Eleni Nisioti, Sebastian Risi. Published under a Creative Commons Attribution 4.0 International (CC BY 4.0) license.}


\section{Introduction}
Modern computational systems are capable of impressive fault tolerance, using mechanisms such as Error Correcting Codes (ECC), modular redundancy, and sophisticated fallback protocols to ensure graceful degradation \citep{shannon_probabilistic_1956, woods_robustness_2008}. However, these engineered successes typically rely on \textit{anticipation}: resources are pre-allocated and failure modes are modeled in advance. When a system encounters damage that exceeds its redundancy budget or defies its failure model, it often fails brittly.

\begin{figure*}[t]
    \centering
    \includegraphics[width=\textwidth]{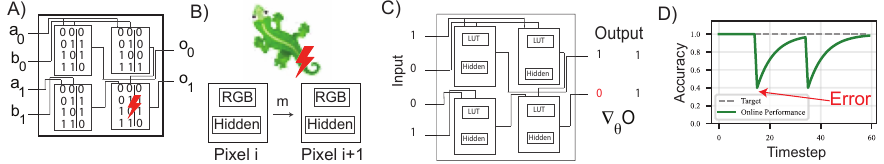}
    \caption{\textbf{From Self-Organising Images to Self-Organising Circuits.} A)~A digital circuit of LUT-based gates can be corrupted by errors that flip LUT contents. B)~NCAs achieve self-organising, self-repairing images via local message passing~\citep{mordvintsev_growing_2020}. C)~We extend this paradigm to circuits: each LUT maintains a hidden state and updates its logic through neighbour communication. D)~At deployment, the trained local policy repairs damage without global supervision.}
    \label{fig:NCA_vs_SODC}
\end{figure*}

Biological intelligence offers a complementary paradigm: \textit{adaptive plasticity}. Natural automata, such as the mammalian cortex, do not rely solely on static backups. Instead, they maintain function by dynamically re-purposing surviving components to compensate for injury, a phenomenon known as cortical remapping \citep{nudo_recovery_2013} (The \textit{capacity} for such plasticity is itself the product of a long prior optimisation process: evolution in biology, and meta-training in our system.) It is enabled by the \textit{degeneracy} of biological networks: the availability of structurally distinct yet functionally equivalent pathways \citep{edelman_degeneracy_2001}.

Inspired by these principles, and by von Neumann's early postulate that systems should ``operate across errors'' \citep{von_neumann_theory_1966}, we aim to provide digital hardware with similar capabilities. As emphasised by Woods et al.\ \citep{woods_robustness_2008}, true hardware autonomy is essential for remote deployments such as deep-space missions out of contact with Earth. This requires moving beyond simple redundancy toward organism-like self-healing, where a collective maintains function despite the loss of individual components. However, in current reconfigurable computing systems like Field-Programmable Gate Arrays (FPGAs), this error mitigation is limited by the `reconfiguration penalty': the overhead of globally monitoring faults and remapping logic \citep{woods_robustness_2008}.

To address this, we introduce a framework for self-organising digital circuits that replaces global, supervised reconfiguration with a decentralised local policy. We extend the Neural Cellular Automata (NCA) paradigm from pattern formation to functional logic generation. In an NCA, simple agents communicate only with their local neighbours, yet collectively produce globally coordinated behaviour. We apply this principle to a substrate of programmable logic gates, where each agent corresponds to a gate whose truth table is a differentiable parameter. When a gate fails, either due to being reversibly corrupted or permanently damaged, the circuit must be reconfigured. The goal of reconfiguration is to recover the original input-output function (\cref{fig:NCA_vs_SODC}).

Our architecture is meta-learned in two phases. The local update rule is trained offline via backpropagation through time, exploiting the differentiability of the continuous LUT relaxation. Once trained, it optimises circuits using only local forward passes, so the expensive global optimisation is paid once and the resulting policy can assemble and repair circuits without differentiable hardware components \citep{sunada_blending_2025}.

\subsection{Key Contributions}
\begin{itemize}
\item We introduce the Topology-Masked Transformer (TMT), a shared-weight attention block applied recurrently under a binary wiring mask, amortising circuit optimisation into a deployment-time forward pass with no differentiable hardware required.
\item The TMT achieves >99.99\% recovery on soft-error patterns up to 5$\times$ larger than seen in training, with an edit size independent of perturbation size.
\item Policies trained on random topologies generalise to unseen larger circuits of 1.7x the training width (264 $\rightarrow$ 450+ nodes).
\end{itemize}

\section{Background and Related Work}
\label{section:Background}

\textbf{Evolvable Hardware.} Evolvable hardware (EHW) applies evolutionary algorithms to discover or repair circuit configurations on reconfigurable substrates such as FPGAs \citep{thompson_evolved_1997, whitley_resurrecting_2021}. EHW has demonstrated circuit evolution and online fault recovery, sharing our core objective of autonomous hardware adaptation. However, evolutionary search incurs its full computational cost for each individual circuit: chromosome lengths grow with circuit complexity, and the resulting search spaces demand substantial resources even for moderately sized designs \citep{haddow_challenges_2011}. Our framework addresses this by amortising the cost of optimisation across all circuits seen during training, replacing stochastic population-based search with a single deterministic forward pass at deployment time. 


\textbf{From Pattern Formation to Functional Substrates.} Neural Cellular Automata (NCA) parameterise local update rules with neural networks, enabling self-organising pattern formation on grids \citep{mordvintsev_growing_2020}. Recent extensions have generalised NCAs to arbitrary graph topologies \citep{grattarola_learning_2021} and to dynamic computational tasks \citep{bena_path_2025}, but the automaton's state always remains the data itself. 

We propose a fundamental shift: viewing the cellular state as the \textit{functional substrate} that processes data. Each cell becomes a programmable logic gate whose truth table is optimised by a learned local policy, rendering the circuit a self-organising system. Because the gates converge to discrete Boolean logic, the resulting circuits remain amenable to formal verification \citep{kresse_logic_2025}.

Concurrently, Miotti et al.\ \citep{miotti_differentiable_2025} use differentiable logic gates to parameterise NCA update rules for hardware-native implementations; our goal is orthogonal, as we use an NCA-like model to optimise the gates themselves.

\textbf{Attention-Based Local Policies.} Our update rule is a deliberate choice of graph neural network (GNN). Rather than the fixed, content-independent neighbour aggregation of convolutional message-passing networks, we use attention: Graph Attention Networks \citep{velickovic_graph_2018} introduced learnable, content-dependent message weighting on graphs, and our policy is a graph attention network specialised to a wired Boolean circuit. The policy must behave like a local \emph{optimiser}: at different stages of assembly and repair, and for different fault patterns, a gate must weigh messages from its neighbours differently, conditioned on their current logits and error. Content-dependent attention supplies this gating, which a fixed-kernel convolutional aggregator cannot. We adopt the Transformer specifically for two properties: (i) weights are shared across all nodes and independent of circuit size, giving scale-freedom; (ii) Pre-LN, QK-normalisation and ReZero keep the block stable under the many recurrent applications the policy requires. Meanwhile,  the binary topology mask enforces a strict, NCA-style one-hop locality. We keep the design deliberately minimal: a single shared-weight block with no structural or edge encodings beyond the wiring mask, so the policy must internalise routing from topology alone; richer Graph Transformer encodings \citep{dwivedi_generalization_2021} are a direction we leave open. Applied recurrently, this topology-masked Transformer (TMT) must learn a local policy that replaces explicit global optimisers, connecting to recent work on meta-trained in-context learners \citep{kirsch_meta-learning_2022, kirsch_general-purpose_2024}.

\section{Self-Organising Digital Circuits Approach}


We model a digital circuit as a Directed Acyclic Graph (DAG) of programmable Look-Up Tables (LUTs) connected by fixed wires (see \cref{fig:NCA_vs_SODC}A). The circuit receives a global binary input $\mathbf{x} \in \{0,1\}^{N_{in}}$ and produces an output $\hat{\mathbf{y}} \in \{0,1\}^{N_{out}}$. Each gate has arity $k$ and is parameterised by a LUT of $2^k$ entries that fully specify its Boolean function. Wires are integer indices selecting bits from the previous layer's outputs, allowing arbitrary connectivity and fan-out. Gates are arranged in feed-forward \emph{circuit} layers (distinct from the internal layers of the Transformer policy introduced below); for our 12-bit tasks with $k=4$, this yields a three-hidden-layer architecture of sizes $(96, 96, 48)$.

During deployment, two categories of hardware fault may corrupt a gate's LUT: (i) \emph{recoverable (soft) errors}, which flip LUT entries but can be overwritten, and (ii) \emph{permanent (stuck-at) faults}, which clamp a gate's output irreversibly. Given a target Boolean function $f$, the goal is to learn a \textit{decentralised local policy} that configures the LUTs such that $\hat{\mathbf{y}} = f(\mathbf{x})$, and that can autonomously restore this mapping after previously unseen faults, without global supervision or backpropagation at deployment time.

Circuits are initialised as noisy ``soft wires'': each gate's LUT is set so the gate merely relays one of its inputs (an identity pass-through, assigned round-robin), with small additive noise to break symmetry. We call these \emph{soft} because the LUT entries remain continuous and uncommitted to any discrete Boolean function; the policy's task is to differentiate them into the target logic.

\subsection{Model: Topology-Masked Transformer (TMT)}
\label{sec:architecture}

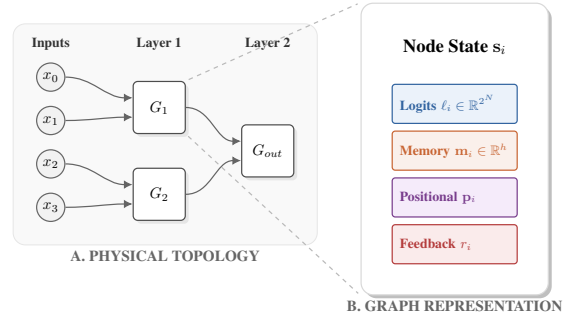
\begin{figure}[t!]
    \centering
    \resizebox{0.95\linewidth}{!}{%
        \input{figures/circuit_schematic_2.tikz}
    }
    \caption[Circuit Architecture and Graph Representation]{\textbf{Circuit Architecture and Graph Representation.} Circuit Architecture and Graph Representation. (A) The physical topology of the circuit. Gates (nodes) are arranged in layers and connected by wires (edges). (B) The graph representation of a single node state $\mathbf{s}_i$, concatenating the functional parameters with the variables that the TMT uses to optimise the circuit. The functional parameter is the LUT logit vector $\ell_i \in \mathbb{R}^{2^N}$, a differentiable lookup table. The hidden variables are the recurrent memory $\mathbf{m}_i \in \mathbb{R}^h$ (the optimiser's hidden state), the positional encoding $\mathbf{p}_i$ (sinusoidal depth awareness, $\sin (l / L)$ ), and the error feedback $r_i$ (a local error signal, defined on output nodes only).}
    \label{fig:circuit_schematic}
\end{figure}

To apply self-organising principles, we lift the circuit into a graph $\mathcal{G} = (\mathcal{V}, \mathcal{E})$ where each node $v_i$ corresponds to a gate or input pin (\cref{fig:NCA_vs_SODC}C). We then learn a local message-passing policy that discovers communication protocols to satisfy the target logic; \cref{fig:circuit_schematic} illustrates the circuit topology and the per-node state representation.

\textbf{Node State.}
Each node carries a state vector: $$\mathbf{s}_i = [\boldsymbol{\ell}_i, \mathbf{m}_i, \mathbf{p}_i].$$
The LUT logits $\boldsymbol{\ell}_i \in \mathbb{R}^{2^k}$ defining the gate's logic, a latent memory $\mathbf{m}_i \in \mathbb{R}^{d_{hidden}}$ ($d_{hidden}=64$) for recurrent state, and a sinusoidal positional encoding $\mathbf{p}_i$ of the normalised depth $d_i/D$, where $d_i$ is the layer index of node $i$ and $D$ the total number of circuit layers. Normalising the depth by $D$ keeps this encoding invariant to circuit scale. Optionally, a scalar feedback signal $r_i$ (described below) is appended, yielding $\mathbf{s}_i = [\boldsymbol{\ell}_i, \mathbf{m}_i, \mathbf{p}_i, r_i]$.

\textbf{Connectivity.}
The graph topology mirrors the circuit wiring. In the random-topology regime, connections are generated by randomly permuting previous-layer output indices, ensuring uniform fan-out while preserving the DAG structure. We enforce bidirectional edges: every forward wire $A \rightarrow B$ implies a backward message-passing edge $B \rightarrow A$.

\textbf{Update Rule.}
We parameterise the update rule as a single-block Transformer operating on the $N$ gate tokens simultaneously. We write $\hat{\mathbf{M}}$ for a layer-normalised \citep{ba_layer_2016} matrix $\mathbf{M}$. The node states $\mathbf{S}$ are projected into a latent space ($d_{attn}=128$):
\begin{equation}
    \mathbf{Z}^{(0)}  = \hat{\mathbf{S}}\, W_{in}^T \in \mathbb{R}^{N \times d_{attn}}
\end{equation}
A single block of masked multi-head self-attention (MSA), whose attention is restricted to wired neighbours via a binary topology mask $M \in \{0,1\}^{N \times N}$, refines the latents. We adopt Pre-LN normalisation \citep{xiong_layer_2020} with separate LayerNorms for queries and keys/values, and QK-normalisation \citep{dehghani_scaling_2023} to stabilise attention logits across recurrent steps. Residual branches are gated via ReZero \citep{bachlechner_rezero_2020}: learned scalars $\alpha$ initialised at zero so that the block initially acts as an identity:
\begin{align}
    \mathbf{Z}'      & = \mathbf{Z}^{(0)} + \alpha_{\text{attn}} \cdot \text{MSA}\!\left(\hat{\mathbf{Z}}^{(0)},\; \hat{\mathbf{Z}}^{(0)},\; M\right) \\
    \mathbf{Z}_{out} & = \mathbf{Z}' + \alpha_{\text{ffn}} \cdot \text{MLP}\!\left(\hat{\mathbf{Z}}'\right)
\end{align}
The two-layer MLP (with GeLU activation) acts independently per node; all cross-node communication is confined to the masked attention step. Output latents are decoded into residual parameter updates via $\alpha$-gated linear heads:
\begin{align}
    \Delta \boldsymbol{\ell}_i & = \alpha_{\ell} \cdot \hat{\mathbf{Z}}_{out,i}\, W_{\ell}^T, &
    \Delta \mathbf{m}_i        & = \alpha_{m} \cdot \hat{\mathbf{Z}}_{out,i}\, W_{m}^T
\end{align}
Positional encodings $\mathbf{p}_i$ remain static; the error feedback $r_i$ is dynamically recomputed at every step. The updates are applied residually ($\mathbf{s}_i^{(t+1)} = \mathbf{s}_i^{(t)} + \Delta\mathbf{s}_i^{(t)}$) by applying this single shared-weight block recurrently for $T$ steps, forming a $T$-layer weight-tied residual network whose expressivity comes through iterated refinement. The topology mask imposes a strict ``speed of light'': information propagates at most one hop per step, so global coordination emerges from iterated local interactions, faithful to the NCA paradigm.

\textbf{Per-Node Error Feedback.}
For fixed-wiring experiments the architecture above suffices; for random wirings, we found explicit error signals essential. Each output gate receives a scalar $r_i$: the mean absolute residual over a task batch:
\begin{equation}
    r_i = \frac{1}{|\mathcal{D}|}\sum_{(\mathbf{x}, \mathbf{y}) \in \mathcal{D}} |y_i - \hat{y}_i(\mathbf{x})|
\end{equation}
Non-output gates receive $r_i = 0$. This signal propagates upstream through recurrent attention, enabling interior gates to adjust in response to downstream error.

\textbf{Scale-Free Architecture.}
Because the Transformer operates at the node level with shared weights, its parameters are independent of circuit size: the same update rule applies to 20 or 200 gates, 3 or 10 layers. Only the topology mask $M$ must be recomputed from the wiring. Combined with normalised positional encodings, this gives the architecture \textit{scale-freedom}: a trained policy can, in principle, be deployed on circuits of different size without retraining.

\subsection{Training}
\label{subsec:training}

\textbf{Differentiable Circuit Execution.}
To train the update rule via backpropagation, we require gradients through the circuit's Boolean logic. Following the differentiable logic gate network (DLGN) paradigm \citep{petersen_deep_2022}, we achieve this through a continuous relaxation: binary signals are represented as probabilities $x \in [0,1]$, and each gate's $2^k$ LUT logits are passed through a sigmoid. 

Unlike the categorical relaxation of DLGNs, which learns a softmax distribution over the 16 fixed two-input Boolean functions, our formulation directly parameterises the continuous LUT entries. The soft gate output is then computed as a multilinear interpolation of the LUT at the input point $\mathbf{x}$. 
For instance, with a 2-input gate, the first input $x_1$ interpolates between the two halves of the LUT to produce an intermediate tensor $L'$:
$$L^{\prime}=\left(1-x_1\right) \cdot\left[l_{00}, l_{01}\right]+ x_1 \cdot\left[l_{10}, l_{11}\right]$$
The second input $x_2$ does the same in $L^{\prime}$ and computes the final scalar output: 
$$y =\left(1-x_2\right)\cdot L^{\prime}[0]+x_2 \cdot L^{\prime}[1].$$ 
This interpolation is fully differentiable, and because each LUT entry influences multiple input combinations, gradients flow through the entire function. At inference, rounding to $\{0,1\}$ recovers discrete Boolean logic.


\textbf{Pool-Based Meta-Learning.}
We frame the circuit optimisation problem as a dynamic system trained via Backpropagation Through Time (BPTT). This is meta-learning in the sense of \citet{andrychowicz_learning_2016}: The TMT is trained to produce parameters for a \textit{separate} computation (the circuit), with the outer objective being that computation's task loss, a two-level structure absent in standard NCAs, where the cell state \textit{is} the target output. As a baseline we compare against standard Backpropagation (BP): gradient descent (Adam) applied directly to all LUT logits over the same continuous relaxation and task loss. Unlike the TMT's local, masked, gradient-free forward passes, BP optimises the whole circuit jointly with global gradient information, so it serves as a centralised, full-information upper bound that the decentralised policy must approximate. Following \cite{mordvintsev_growing_2020}, we maintain a persistent graph Pool $\mathcal{P} = \{ \mathcal{G}_1, \dots, \mathcal{G}_K \}$. Circuits are periodically reset to unoptimised states (average lifespan of 128 steps) to enforce a dual curriculum: constant injection of fresh circuits demands rapid self-assembly, while surviving circuits enforce long-term homeostasis under damage injections.

\textbf{Inner Loop (Inference):} At each training step, a batch of circuits and Boolean input--output pairs $(\mathbf{X_{train}}, \mathbf{Y_{train}})$ are sampled. The TMT is applied iteratively for $T$ steps. Crucially, the circuit is functionally executed at every tick $t$: updated LUT logits are extracted, the circuit processes the input data, and resulting per-node residuals $r_i$ are written back into the graph state. This creates a closed real-time feedback loop. To balance recurrent expressivity with the computational constraints of scaling batch size and model capacity, we truncate the BPTT horizon to $T=5$.

\textbf{Outer Loop (Optimisation):} Following the $T$-step scan, the circuit loss $\mathcal{L}$ (Binary Cross Entropy) is computed against $\mathbf{Y_{train}}$. We support both fixed evaluation at the final step ($t^{*} = T$) and stochastic evaluation ($t^{*} \sim \mathcal{U}(T_{\min}, T)$) to capture varying gradient depths and enable curriculum scheduling. Gradients of $\mathcal{L}$ with respect to the TMT parameters are computed via BPTT through both the recurrent message-passing and the functional circuit execution, followed by an Adam \citep{kingma_adam_2017} update.

\section{Experiments and Results}
\label{sec:experiments}

\textbf{Boolean Tasks.}
We evaluate on three 12-bit tasks, each enumerating all $2^{12}=4{,}096$ input--output pairs; $256$ ($6.25\%$) are held out as a test set by a per-run random permutation of the inputs. As each task is a bijection, no pairs are duplicated or mirrored, train and test indices are disjoint (no leakage), and no inputs are excluded or stratified (the all-zero and other trivial cases may fall in either split). \textbf{Split Multiplication:} two 6-bit integers are multiplied, occupying the full 12-bit output. \textbf{Split Addition:} two 6-bit integers are added (7-bit result, 5 bits zero-padded). \textbf{Bit Reversal:} the input array is reversed, mapping bit $i$ to position $11-i$. These three tasks are deliberately minimal, exactly-specified Boolean functions spanning a difficulty gradient: from pure routing (Bit Reversal, a permutation) through carry-dependent arithmetic (Addition, then the harder Multiplication). We use them as interpretable, controlled probes of the local policy rather than as an end application; the framework itself is task-agnostic.

\textbf{Metrics.} Throughout, \emph{soft} accuracy and loss are measured on the continuous (sigmoid-relaxed) circuit outputs, whereas \emph{hard} accuracy is the fraction of output bits matched exactly after rounding both the LUTs and the outputs to $\{0,1\}$, i.e.\ the true discrete Boolean behaviour of the deployed circuit. \emph{Final} hard accuracy is measurement at the end of the policy's deployment rollout (256 steps; for BP, at convergence). Unless noted otherwise, reported numbers are final hard accuracies on the held-out test split.

\begin{figure}
    \centering
    \includegraphics[width=\linewidth]{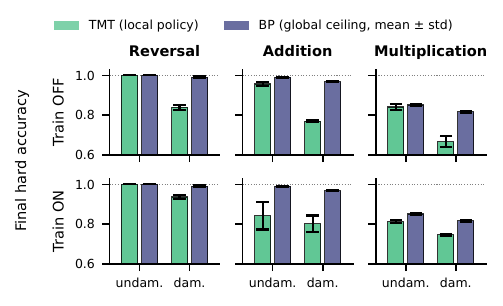}
    \caption[TMT vs.\ BP on Fixed Topologies.]{\textbf{TMT vs.\ BP on fixed topologies.} Final hard accuracy on the held-out test split after 256 steps, as \textbf{TMT} (green) and \textbf{BP-ceiling} (indigo) bars (mean$\pm$std over 5 seeds). In each panel the left pair is evaluated without damage, the right pair under stochastic damage (clamping 20\% of gates). \textbf{Rows:} policy trained without (top) vs.\ with (bottom) damage; \textbf{columns:} task. Undamaged, TMT matches the BP ceiling; under damage it degrades but damage-aware training (bottom row) recovers most of the gap.}
    \label{fig:fixed_wiring}
\end{figure}


\subsection{Regime I: Growth, Persistence, and Repair on Fixed Topologies}
\label{subsec:regime1}

\begin{figure}[ht!]
    \centering
    \includegraphics[width=\linewidth]{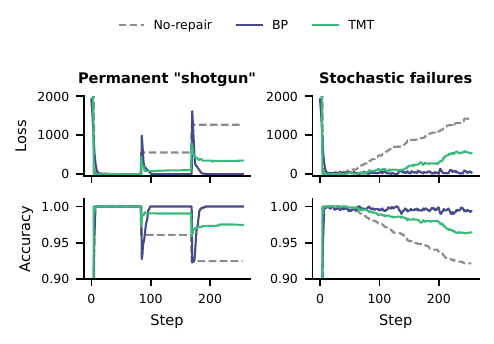}
    \caption[TMT Resilience to Damages.]{\textbf{Resilience to permanent damage.} Step-wise loss (top) and hard accuracy (bottom) under permanent ``shotgun'' knockouts (left) and stochastic failures (right). \textbf{TMT} (green) re-repairs after each event toward the \textbf{BP} ceiling (indigo); the no-repair baseline (grey dashed) decays.}
    \label{fig:resilience}
\end{figure}

We first establish that the local TMT policy can grow functional circuits from unoptimised ``soft wires'' and maintain them indefinitely.

As shown in \cref{fig:fixed_wiring} (top row, undamaged eval), the TMT policy converges across all tasks without damage, achieving performance virtually identical to the global Backpropagation (BP) baseline. It achieves perfect accuracy on Bit Reversal and mean accuracies of 0.96 and 0.84 on Split Addition and Multiplication, respectively.

To evaluate fault tolerance, we introduce stochastic damage (clamping 20\% of gates to zero). When exposed to damage out-of-distribution (Zero-Shot Resilience, \cref{fig:fixed_wiring}, top row, damaged eval), the TMT retains partial functionality, indicating inherent robustness in the distributed representation. When trained with active damage (Learned Resilience, Bottom Row), performance under damage approaches the BP baseline, demonstrating adaptive robustness, albeit with a slight degradation in the damage-free ceiling due to the noisy training environment.

Targeted catastrophic events introduce simultaneous destruction of 10\% of its gates, termed ``shotgun'' in \cref{fig:resilience}. The TMT actively recovers and stabilises at $\approx 0.97$ accuracy, vastly outperforming a passive (no-repair) baseline. While global BP (upper bound) recovers more fully, the TMT's decentralised repair minimises the initial impact drop.

\begin{figure}[h]
    \centering
    \includegraphics[width=\linewidth]{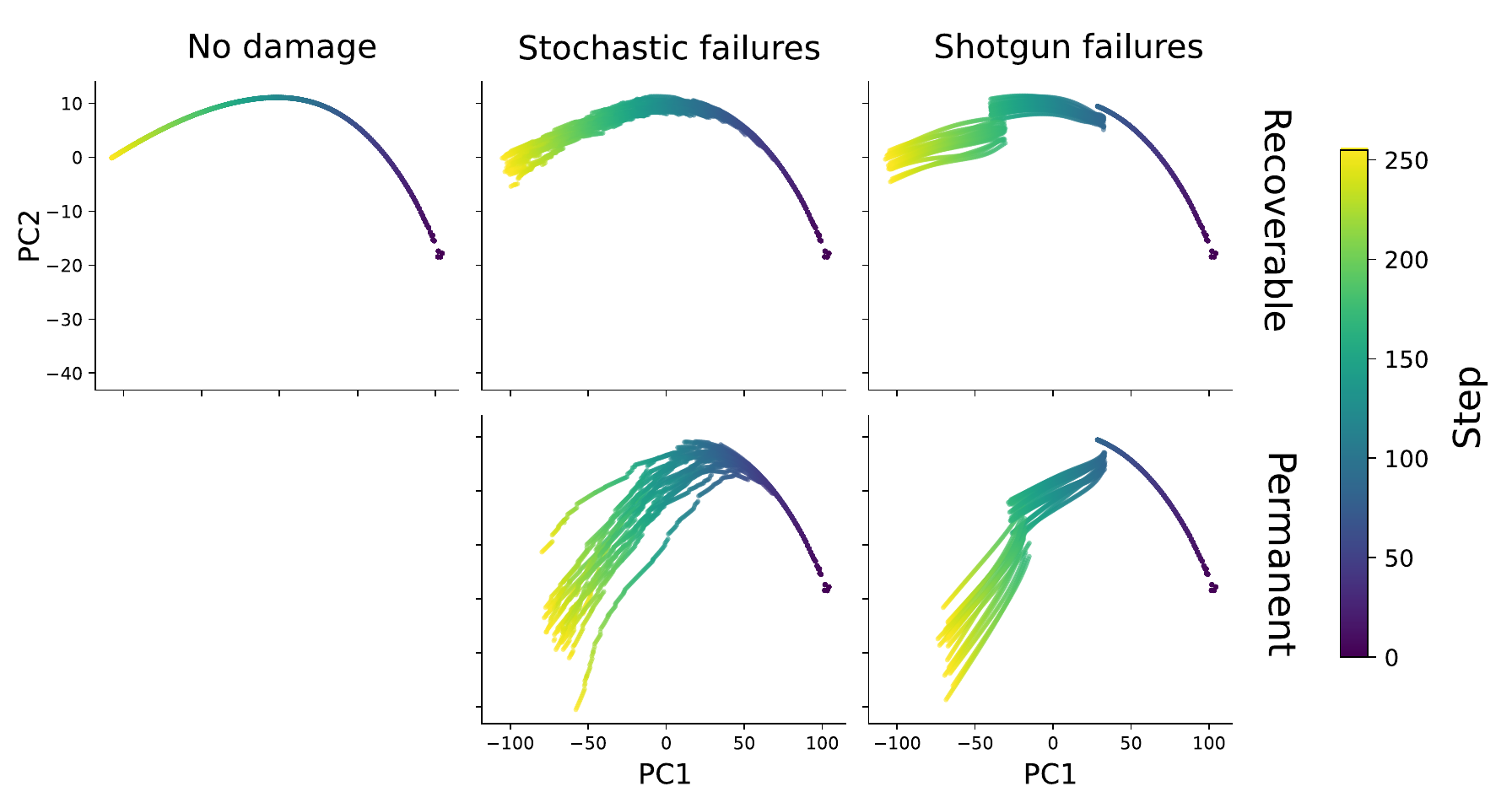}
    \caption{\textbf{PCA Trajectories of Circuit Optimisation.} Each trajectory is the path of a circuit's LUT-logit configuration vector over optimisation steps, projected onto its first two principal components (colour denotes step). \textbf{(Left)} Without damage, optimisation is near-deterministic. \textbf{(Middle/Right)} Under recoverable or permanent damage, trajectories diverge, driving circuits into a diverse fan of alternative configurations rather than rewinding. Axes are the leading linear components and should be read qualitatively.}
    \label{fig:pca_trajectories}
\end{figure}

Projecting the circuits' LUT-logit configuration vectors along optimisation onto their first two principal components (\cref{fig:pca_trajectories}) illustrates this resilience: under recoverable damage, the TMT does not rewind to a single canonical state but dynamically re-routes, fanning out into functionally equivalent yet structurally distinct configurations. As a linear projection, PCA captures only the dominant axes of variation; we use it here to preserve the global geometry of these trajectories, and reserve the non-linear UMAP of \cref{fig:50k_umap} for resolving cluster structure in the much larger recovered solution set.

\subsection{Regime II: Self-Healing Circuits and the Degenerate Solution Space}
\label{subsec:regime2}
To focus on practical considerations, we remove the need to discover functioning circuits and test the TMT purely as a maintenance mechanism on circuits preconfigured to perfect accuracy by BP. The two regimes probe complementary fault types by design: Regime~I examined \emph{permanent} (stuck-at) faults - the harder case, in which damaged gates cannot be reused and the policy must re-route logic around them, whereas here we study \emph{reversible} (soft) errors.

\begin{figure}[t]
    \centering
    \includegraphics[width=0.75\linewidth]{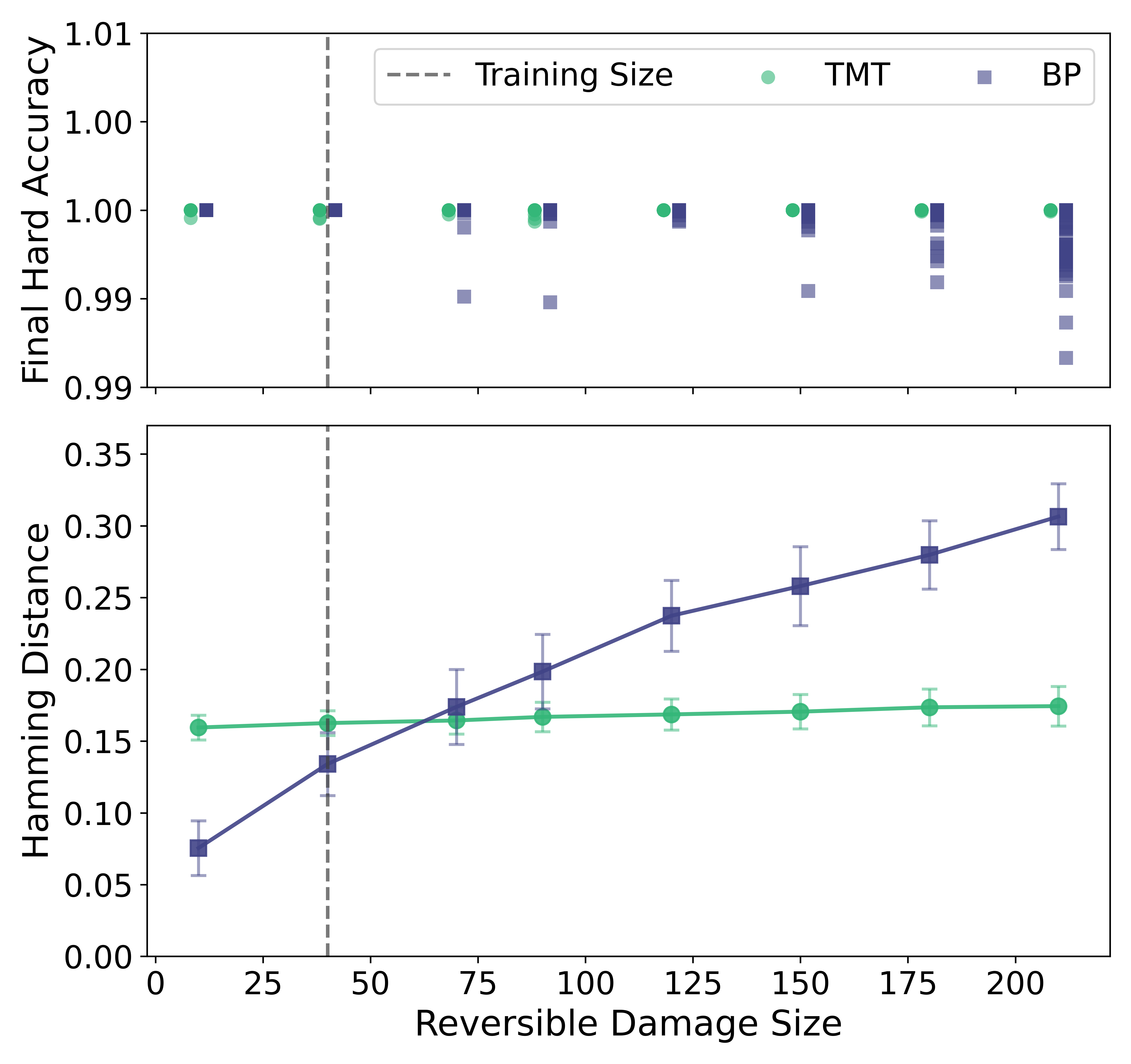}
    \caption{\textbf{Out-of-Distribution Perturbation Recovery.} \textbf{(Top)} Hard accuracy of recovered solutions from out-of-distribution damage patterns of increasing size, after training on 40-gate patterns only (dashed v-line). 100 patterns per damage size are shown (20 random patterns across 5 seeds).  \textbf{(Bottom)}   Mean per-gate Hamming distance between base and recovered circuits.}
    \label{fig:OOD_reversible}
\end{figure}

These reversible perturbations model the radiation-induced soft errors that dominate over hard faults in aerospace-grade SRAM-based FPGAs \citep{wang_review_2024}. \Cref{fig:OOD_reversible} compares TMT and BP performance upon damage recovery to convergence (26 message steps and 300 gradient steps, respectively), on the \textbf{Binary Addition} task. For TMT, it demonstrates near-perfect out-of-distribution recovery for "shotgun" damage sizes far exceeding training parameters, achieving >99.99\% hard accuracy across all perturbations. This generalisation highlights the model's relevance for radiation-induced error scenarios. Hamming distances reveal a ``signature edit'' fraction independent of perturbation size, contrasting with the monotonically increasing response exhibited by BP.

\begin{figure}
    \centering
    \includegraphics[width=0.9\linewidth]{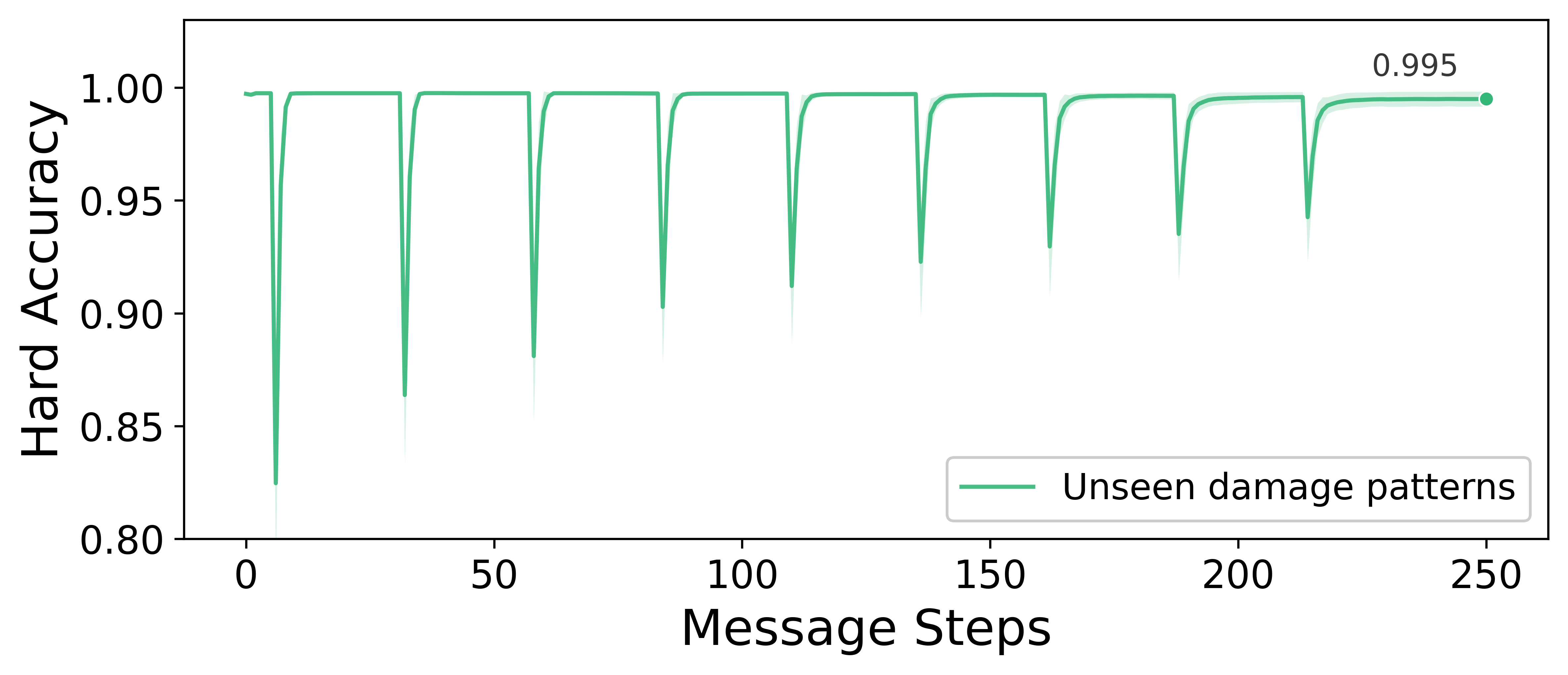}
    \caption{\textbf{Recovery Trajectory from Reversible Damage.} Recovery from successive 40-gate damage events; at each event the 40 gates are drawn independently and uniformly at random from the eligible hidden gates (a fresh draw per event, not a fixed set). The shaded region is the standard deviation across 256 such random damage patterns.}
    \label{fig:multidamage}
\end{figure}

To further stress-test the system, we introduce successive "shotgun" failures of 40 gates at once (17\%). \Cref{fig:multidamage} demonstrates recovery on the addition task following multiple damage events.

\begin{figure}[t!]
    \centering
    \includegraphics[width=\columnwidth]{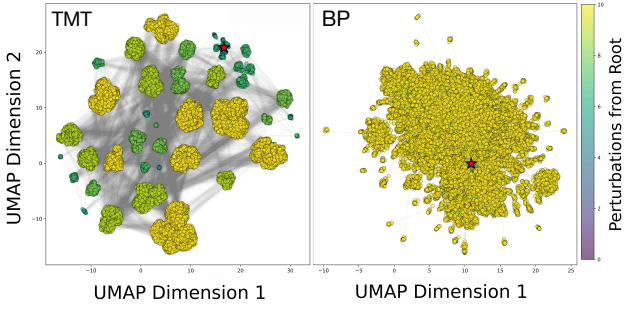}
    \caption{\textbf{UMAP of Circuit Solution Space.} 2D embedding of 50,000 functional configurations recovered by the TMT policy (Top) and BP (Bottom) computing binary addition. The red star indicates root circuit, grey edges indicate perturbation - recovery cycles. The TMT policy discovers discrete structural archetypes while BP showcases a single cluster.}
    \label{fig:50k_umap}
\end{figure}

\textbf{Solution Degeneracy.}
To map the degenerate solution space of these self-healing circuits (\cref{fig:50k_umap}), we employ a recursive Depth-First Search (DFS) exploration strategy. Starting from a preconfigured circuit with perfect accuracy on the binary addition task (8 inputs/outputs, yielding a 2560-dimensional LUT vector), we apply a 40-gate recoverable perturbation. Because the damage is not permanent, the exact pre-damage state is theoretically recoverable. However, the system consistently finds alternate, functionally equivalent solutions (perfect accuracy).

By treating each recovered circuit as a new seed for subsequent perturbations, we generate a recursive tree of trajectories (search depth of 10, branching factor of 4). We apply this procedure identically to the TMT policy and a standard BP baseline, visualising the resulting 2560-dimensional configuration space via UMAP (Euclidean distance, $n_{\text{neighbors}}=15$, $\text{min\_dist}=0.1$), where Euclidean distance preserves the neighbourhood ranking of Hamming distance for binary vectors.

The TMT policy organises recovered solutions into discrete functional clusters. The sparse connectivity between these clusters suggests a \textit{neutral landscape}: multiple successive perturbations force the system to traverse fitness valleys, landing in entirely different structural arrangements that maintain identical global functionality. (UMAP distances are non-metric, so the embedding is best read qualitatively; the discreteness we highlight is corroborated by the perturbation--recovery graph itself, whose inter-cluster edges are sparse.) This is consistent with the policy exploring a broad, degenerate solution space rather than returning to a single basin.

In contrast, standard BP produces a single cluster centered on the root configuration, exploring more incremental deformations rather than traversing between distinct structural basins. 

\subsection{Regime III: Generalisation to Random Topologies}
\label{subsec:regime3}

The most challenging setting removes the architectural constraints entirely. In the Random Topology regime, every circuit in the pool possesses a unique, randomly generated wiring diagram. Consequently, the meta-learner cannot overfit to ``Gate A connected to Gate B.'' Instead, it must learn a truly topological, wiring-agnostic policy.

\begin{figure}[t]
    \centering
    \includegraphics[width=\columnwidth]{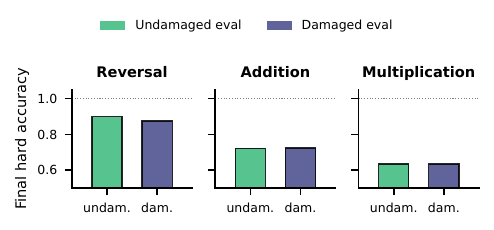}
    \caption[TMT on Random Topologies.]{\textbf{TMT on random topologies.} Final hard accuracy on held-out random wirings after 256 steps, evaluated without vs.\ with stochastic damage (single seed). The BP baseline (inherently wiring-agnostic) matches \cref{fig:fixed_wiring}.}
    \label{fig:random_wiring}
\end{figure}

We find this learning regime to be significantly harder than Fixed Topology training. We successfully learned a high-performing, topology-agnostic policy for the \textbf{Bit Reversal} task, achieving good generalisation to unseen random graphs. However, for the arithmetic tasks (Addition and Multiplication), the policy struggles to fully converge. While it learns to approximate the output distribution, capturing coarse statistical patterns of the target function, it lacks the precision required for exact arithmetic operations.
In this specific regime, standard BP (which is inherently topology-agnostic via re-training) still holds the advantage. However, the success on the Bit Reversal task provides a proof-of-principle that the TMT \textit{can} learn generalisable routing algorithms. We display random-wiring training results in \cref{fig:random_wiring}.

\subsection{Regime IV: Scale-Free Optimisation}
\label{subsec:regime4}

Finally, we investigate the scaling capability of the TMT policy on circuits significantly larger (or smaller) than those seen during training. 

\begin{figure}[t]
    \centering
    \includegraphics[width=\columnwidth]{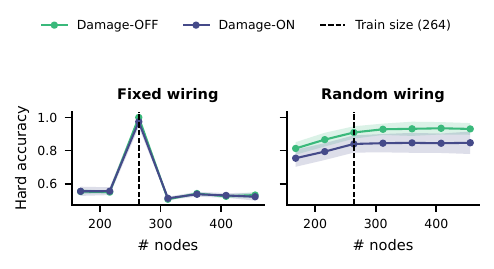}
    \caption[Scale-Free Generalisation]{\textbf{Scale-Free Generalisation.} Test hard accuracy vs.\ circuit size (number of nodes); dashed line marks the training size ($N=264$). \textbf{Fixed} training (left) overfits scale: accuracy peaks at the training size and collapses elsewhere,whereas \textbf{random} training (right) generalises, even \textit{improving} on wider circuits.}
    \label{fig:scale_fixed}
    \label{fig:scale_random}
\end{figure}

The results, displayed in \cref{fig:scale_random}, reveal a critical insight: scale-freedom is not just an architectural feature, but rather a \textit{learned capability}.

\textbf{Overfitting Size (Fixed Training):} When the TMT is trained on a fixed topology (\cref{fig:scale_fixed}), it overfits the specific scale of the training graph. Performance peaks exactly at the training size ($N=264$, vertical dashed line) and collapses for larger or smaller circuits.

\textbf{Emergent Scalability (Random Training):} In contrast, when trained on the curriculum of random topologies (\cref{fig:scale_random}), the policy generalises remarkably. Not only does it maintain function on larger circuits, but accuracy actually \textit{increases} as we expand the width of the circuit (from 264 to 450+ nodes). The local policy is able to utilise the additional latent capacity of the wider layers to route signals more effectively, despite never having encountered graphs of this size during training.

We note that this scaling success is currently limited to circuit \textit{width}. Scaling \textit{depth} remains a challenge, likely because our positional encoding (normalised depth fraction) changes resolution as layers are added, disrupting the policy's depth perception. Nevertheless, the ability to train on small, cheap circuits and deploy on wider architectures is encouraging evidence for the TMT's ``growth'' paradigm.

\section{Discussion}
In summary, we extend the NCA paradigm from grid-based pattern formation to functional logic generation on arbitrary graphs: replacing global backpropagation with a decentralised, topology-masked Transformer, circuits self-assemble, self-repair, and generalise their routing across structural scales. For soft errors the policy recovers perfectly several times beyond training-scale damage. Rather than rewinding to the recoverable prior state, settles on new functionally equivalent minima, prioritising functional homeostasis over structure.

This use of degenerate solutions mirrors biological multiscale competency \citep{pezzulo_top-down_2016}, where evolved systems maintain robust homeostasis under noisy, uncertain conditions by dynamically switching between structurally distinct but functionally equivalent pathways (e.g, aerobic vs.\ anaerobic metabolic pathways \citep{edelman_degeneracy_2001}). 
While current industrial FPGA applications typically require deterministic logic configurations, the capacity of the TMT to harness structural degeneracy holds promise for autonomous systems facing unpredictable hardware failures in remote environments such as deep space missions.

\subsection{Limitations and Future Directions}
Despite these capabilities, several limitations present immediate avenues for future work. Most fundamentally, our circuits remain small (hundreds of gates) and our tasks simple relative to production FPGAs, which contain thousands to millions of LUTs; the present results are therefore a proof-of-principle, and closing the gap to hardware-scale deployment is the overarching challenge into which the specific limitations below feed. First, our current positional encoding captures only vertical depth within the circuit DAG, stripping the policy of local topological context. Incorporating Random Walk Structural Encodings (RWSE) \citep{rampasek_recipe_2023}, Laplacian eigenvectors \citep{shuman_emerging_2013}, or functional metrics like fan-out centrality would provide the meta-learner with richer structural maps while preserving the architecture's scale-freedom. Second, the per-node error feedback $r_i$ is a coarse scalar residual. The meta-learner remains blind to the actual task data; augmenting the architecture with cross-attention to input-output pairs, akin to Perceiver IO \citep{jaegle_perceiver_2021}, could enable genuine, data-informed in-context reasoning. Finally, although attention is already restricted to each gate's wired neighbours, our implementation gathers them into fixed-width padded neighbourhoods, which wastes computation when gate degrees are uneven and still grows with circuit size; a fully sparse attention kernel will be needed to scale efficiently to circuits with thousands of gates, where physical connectivity is inherently sparse ($<3\%$ density).

\subsection{Toward Self-Organising Computational Substrates}
The true significance of this work lies beyond the Boolean domain on which we have validated it. Recent self-organising systems already let a substrate compute: developmental graph cellular automata and critical neural cellular automata have both been grown into \emph{reservoirs} whose self-organised dynamics are harnessed for computation \citep{barandiaran_growing_2025, pontes_filho_reservoir_2025}. What distinguishes our setting is the relationship between policy and substrate. In reservoir computing the substrate is selected or evolved for generic computational capacity and then read out by a separately trained layer; its configuration is not directed toward a specified function and is largely fixed at deployment. Here, by contrast, the decentralised policy \emph{directly configures and continually repairs} the substrate's own functional parameters to realise a \emph{specified} Boolean function, with no separate readout. The topology mask defines the physical wiring, while the learned attention policy governs the functional logic of connected nodes. These two levels, structural connectivity and functional coupling, are independently addressable: one can remain fixed while the other adapts, or both can evolve on separate timescales. This dissociation is absent in conventional neural networks, where structure and function are collapsed into a single weight matrix.

The separation naturally suggests a richer architecture in which the same shared-weight mechanism first grows a sparse structural scaffold, then governs the functional dynamics within it, with no hard boundary between the two phases. Realising this structural half, endowing the TMT with the ability to add, prune, and rewire, remains the central open challenge. 

By embracing adaptive plasticity over prescriptive redundancy, this work forms the basis for computational substrates that grow, learn, and heal themselves.

\newpage

\section{Acknowledgements} 

We thank Nicolas Bessone, Ismail Ceylan, Matthias Dellago, Benedikt Hartl, Robin Hiesinger, Kathrin Korte, Milton Montero, Elias Najarro, Joachim W.\ Pedersen, Fernando Rosas, and Florian Scheidl for fruitful and inspiring discussions.

Funded by the European Union (ERC, GROW-AI, 101045094). Views and opinions expressed are however those of the authors only and do not necessarily reflect those of the European Union or the European Research Council. Neither the European Union nor the granting authority can be held responsible for them.

\footnotesize
\bibliographystyle{apalike}
\bibliography{MetaCircuits}

\end{document}

%% file: figures/circuit_schematic_2.tikz

\definecolor{coreBlue}{RGB}{50, 100, 160}
\definecolor{softBlue}{RGB}{235, 242, 250}
\definecolor{coreOran}{RGB}{200, 100, 50}
\definecolor{softOran}{RGB}{250, 240, 235}
\definecolor{corePurp}{RGB}{120, 60, 140}
\definecolor{softPurp}{RGB}{245, 235, 250}
\definecolor{coreRed}{RGB}{180, 60, 60}
\definecolor{softRed}{RGB}{250, 235, 235}
\definecolor{wireGray}{RGB}{100, 100, 100}

\ifdefined\boxw\else\newlength{\boxw}\fi
\settowidth{\boxw}{\small\textbf{Memory} $\mathbf{m}_i \in \mathbb{R}^{h}$}
\addtolength{\boxw}{2pt}

\begin{tikzpicture}[
    font=\sffamily,
    >={LaTeX[width=2mm,length=2mm]},
    gate/.style={
        draw=black!70, 
        thick, 
        fill=white, 
        minimum width=1.2cm, 
        minimum height=1.2cm, 
        rounded corners=3pt,
        drop shadow={opacity=0.15},
        align=center
    },
    input_node/.style={
        circle, 
        draw=black!60, 
        thick, 
        fill=gray!10, 
        inner sep=0pt, 
        minimum size=0.65cm
    },
    wire/.style={
        draw=wireGray, 
        thick, 
        ->
    },
    vec_box/.style={
        rectangle, 
        thick,
        minimum height=0.9cm, 
        text width=\boxw,
        align=left,
        font=\small,
        rounded corners=2pt,
        inner sep=5pt
    },
    column_label/.style={
        font=\bfseries\small, 
        text=black!80, 
        anchor=south
    },
    panel_label/.style={
        font=\bfseries, 
        text=black!60, 
        align=center,
        anchor=north
    }
]


\coordinate (layer_y_top) at (0, 1.2);

\node[input_node] (x0) at (0, 0.5) {$x_0$};
\node[input_node] (x1) at (0, -0.5) {$x_1$};
\node[input_node] (x2) at (0, -1.5) {$x_2$};
\node[input_node] (x3) at (0, -2.5) {$x_3$};

\node[column_label] at (0, 1.0) {Inputs};

\node[gate] (g1) at (2.5, -0.2) {$G_1$};
\node[gate] (g2) at (2.5, -2.2) {$G_2$};

\node[column_label] at (2.5, 1.0) {Layer 1};

\node[gate] (gout) at (5.0, -1.2) {$G_{out}$};

\node[column_label] at (5.0, 1.0) {Layer 2};

\draw[wire] (x0) to[out=0, in=180] (g1.160);
\draw[wire] (x1) to[out=0, in=180] (g1.200);
\draw[wire] (x2) to[out=0, in=180] (g2.160);
\draw[wire] (x3) to[out=0, in=180] (g2.200);
\draw[wire] (g1) to[out=0, in=180] (gout.160);
\draw[wire] (g2) to[out=0, in=180] (gout.200);

\begin{scope}[on background layer]
    \node[fit=(x0)(x3)(gout)(layer_y_top), rounded corners=8pt, fill=gray!5, draw=gray!20, inner sep=15pt] (circuit_bg) {};
    \node[panel_label] at (circuit_bg.south) {A. PHYSICAL TOPOLOGY};
\end{scope}


\coordinate (zoom_top_left) at (8.0, 1.5);

\node[anchor=north west, font=\Large\bfseries] at (zoom_top_left) (title) {Node State $\mathbf{s}_i$};

\node[vec_box, fill=softBlue, draw=coreBlue, text=coreBlue, below=0.55cm of title] (logits) {
    \textbf{Logits} $\ell_i \in \mathbb{R}^{2^N}$
};

\node[vec_box, fill=softOran, draw=coreOran, text=coreOran, below=0.15cm of logits] (memory) {
    \textbf{Memory} $\mathbf{m}_i \in \mathbb{R}^{h}$
};

\node[vec_box, fill=softPurp, draw=corePurp, text=corePurp, below=0.15cm of memory] (pos) {
    \textbf{Positional} $\mathbf{p}_i$
};

\node[vec_box, fill=softRed, draw=coreRed, text=coreRed, below=0.15cm of pos] (err) {
    \textbf{Feedback} $r_i$
};

\begin{scope}[on background layer]
    \node[fit=(title)(logits)(memory)(pos)(err)(zoom_top_left), rounded corners=8pt, fill=white, draw=gray!30, thick, inner sep=20pt, drop shadow] (zoom_bg) {};
    \node[panel_label] at (zoom_bg.south) {B. GRAPH REPRESENTATION};
\end{scope}


\draw[gray!55, dashed, thick, shorten >=2pt, shorten <=2pt] (g1.north east) -- (zoom_bg.north west);
\draw[gray!55, dashed, thick, shorten >=2pt, shorten <=2pt] (g1.south east) -- (zoom_bg.south west);


\end{tikzpicture}